\documentclass[twoside,11pt]{article}
\pdfoutput=1

\usepackage{geometry}
\usepackage{natbib}
\usepackage{graphicx}
\usepackage{rotating}
\usepackage{amsmath}
\usepackage{amssymb}
\usepackage{theorem}
\newcommand{\BlackBox}{\rule{1.5ex}{1.5ex}}  % end of proof

\newcommand{\cbr}[1]{\left\{#1\right\}}

\newcommand{\intset}[1]{\cbr{1..n}}

\usepackage[colorlinks,pdfpagelabels,plainpages=false]{hyperref}
\usepackage{algorithm}
\usepackage{algorithmic}
\usepackage{rotating}

\usepackage{xcolor}
\definecolor{dark-red}{rgb}{0.4,0.15,0.15}
\definecolor{dark-blue}{rgb}{0.15,0.15,0.4}
\definecolor{medium-blue}{rgb}{0,0,0.5}
\hypersetup{
   colorlinks, linkcolor={dark-blue},
   citecolor={dark-blue}, urlcolor={medium-blue}
}

\usepackage{booktabs}
\usepackage{bm}
\usepackage{amsmath}
\usepackage{amssymb}
\usepackage{amsfonts}
\usepackage{mathtools}
\usepackage{comment}
\usepackage{subfigure}

\newcommand{\mbf}[1]{{\boldsymbol{\mathbf{#1}}}}
\renewcommand{\bm}{\mbf}

\usepackage{color}

\usepackage{array}
\newcolumntype{L}[1]{>{\raggedright\let\newline\\\arraybackslash\hspace{0pt}}m{#1}}
\newcolumntype{C}[1]{>{\centering\let\newline\\\arraybackslash\hspace{0pt}}m{#1}}
\newcolumntype{R}[1]{>{\raggedleft\let\newline\\\arraybackslash\hspace{0pt}}m{#1}}

\usepackage{parskip}

\usepackage{multirow}
\usepackage{titlefoot}
\usepackage{gensymb}

\title{Deep Kernel Learning}

\author{
  Andrew Gordon Wilson\footnote{Equal contribution.} \\
  Carnegie Mellon University \\
  \texttt{andrewgw@cs.cmu.edu}
  \and
  Zhiting Hu$^{*}$ \\
  Carnegie Mellon University \\
  \texttt{zhitingh@cs.cmu.edu}
  \and
  Ruslan Salakhutdinov \\
  University of Toronto \\
  \texttt{rsalakhu@cs.toronto.edu}
  \and
  Eric P. Xing \\
  Carnegie Mellon University \\
  \texttt{epxing@cs.cmu.edu}
}

\begin{document}
\date{}

\maketitle

\begin{abstract} 

\begin{sloppypar}
%\normalsize
We introduce scalable deep kernels, which combine the structural properties of deep learning
architectures with the non-parametric flexibility of kernel methods.  Specifically, we transform
the inputs of a spectral mixture base kernel with a deep architecture, using local kernel
interpolation, inducing points, and structure exploiting (Kronecker and Toeplitz) algebra
for a scalable kernel representation.  These closed-form kernels can be
used as drop-in replacements for standard kernels, with benefits in expressive power and scalability.
We jointly learn the properties of these kernels through the marginal likelihood of a Gaussian process.
Inference and learning cost $\mathcal{O}(n)$ for $n$ training points, and predictions cost
$\mathcal{O}(1)$ per test point.  On a large and diverse collection of applications, including a
dataset with $2$ million examples, we show improved performance over
scalable Gaussian processes with flexible kernel learning models, and stand-alone deep architectures.
\end{sloppypar}

\end{abstract}

\section{Introduction}
\label{sec: intro}

``How can Gaussian processes possibly replace neural networks?  Have we thrown the
baby out with the bathwater?'' questioned \citet{MacKay98}.  It was the late 1990s, and researchers had grown frustrated with the many
design choices associated with neural networks -- regarding architecture, activation functions, and regularisation -- and the lack of a principled
framework to guide in these choices.

Gaussian processes had recently been popularised within the machine learning community by
\citet{neal1996}, who had shown that Bayesian neural networks with infinitely many
hidden units converged to Gaussian processes with a particular kernel (covariance) function.
Gaussian processes were subsequently viewed as flexible and interpretable alternatives to
neural networks, with straightforward learning procedures.  Where neural networks used finitely
many highly adaptive basis functions, Gaussian processes typically used infinitely many
fixed basis functions.
As argued by \citet{MacKay98}, \citet{HOT}, and \citet{Bengio2009}, neural networks could automatically discover meaningful representations in high-dimensional
data by
learning multiple layers of highly adaptive basis functions.
By contrast, Gaussian processes with popular kernel functions were
used typically as simple smoothing devices.

Recent approaches~\citep[e.g.,][]{wilson2014thesis, wilsonadams2013, lloyd2014, yang2015carte}
have demonstrated that one
can develop more expressive kernel functions,
which are indeed able to discover rich structure in data without human intervention.
Such methods effectively use infinitely many adaptive basis functions.  The relevant question then
becomes \emph{not} which paradigm (e.g., kernel methods or neural networks) replaces the other, but
whether we can combine the advantages of each approach.  Indeed, deep neural networks provide a
powerful mechanism for creating adaptive basis functions, with inductive biases which have proven
effective for learning in many application domains, including visual object recognition, speech
perception, language understanding, and information retrieval~\citep{alex2012,deepSpeechReviewSPM2012,SocherHPNM11,kirosTACL,Xu2015}.

In this paper, we combine the non-parametric flexibility of kernel methods with
the structural properties of deep neural networks.  In particular, we 
use deep feedforward fully-connected
and convolutional networks, in combination with spectral mixture covariance functions
\citep{wilsonadams2013}, inducing points \citep{quinonero2005unifying}, structure exploiting algebra \citep{saatchi11}, and local 
kernel interpolation \citep{wilsonnickisch2015, wdn2015}, to create scalable expressive closed form covariance kernels for Gaussian processes.
As a non-parametric method, the information capacity of our model grows with the amount of available
data, but its complexity is automatically calibrated through the marginal likelihood of the Gaussian
process, without the need for regularization or cross-validation \citep{rasmussen01, rasmussen06,
wilson2014thesis}.  The flexibility and automatic calibration provided by the non-parametric layer
typically provides a high standard of performance, while reducing the need for extensive hand tuning from
the user.

We further build on the ideas in KISS-GP~\citep{wilsonnickisch2015} and extensions~\citep{wdn2015}, 
so that our deep kernel learning model can scale \emph{linearly} with the number of training instances $n$, instead of
$\mathcal{O}(n^3)$ as is standard with Gaussian processes, while retaining a fully non-parametric
representation. Our approach also scales as $\mathcal{O}(1)$ per test point 
allowing for very fast prediction times.  Because KISS-GP creates an approximate kernel from a user
specified kernel for fast computations, independently of a specific inference procedure, we can view the resulting kernel
as a \emph{scalable deep kernel}.  We demonstrate the value of this scalability in the experimental results section,
where it is the large datasets that provide the greatest opportunities for our model
to discover expressive statistical representations.

We begin by reviewing related work in section \ref{sec: related}, and
providing background
material on Gaussian processes in section \ref{sec: gps}.  In section \ref{sec: dkl} we derive scalable closed form
deep kernels, and describe how to perform efficient automatic learning of these kernels through the Gaussian
process marginal likelihood.  In section \ref{sec: experiments}, we show substantially improved performance
over standard Gaussian processes, expressive kernel learning approaches, and deep neural networks, on
a wide range of datasets. 
We also examine the structure of the kernels
to gain new insights into our modelling problems.

\section{Related Work}
\label{sec: related}
Given the intuitive value of combining kernels and neural networks,
it is encouraging that various
distinct forms of such combinations have been considered in different
contexts.

The Gaussian process regression network \citep{wilson12icml} replaces
all weight connections in a Bayesian neural network with Gaussian
processes, allowing the authors to model input
dependent correlations between multiple tasks.
Alternatively, \citet{damianou2013deep} replace every activation
function in a Bayesian neural network with a Gaussian process transformation, in an
unsupervised setting.  While promising, both models are very task specific,
and require sophisticated approximate Bayesian
inference which is much more demanding than what is required by
standard Gaussian processes or deep learning models, and typically does not
scale beyond a few thousand training points.  Similarly, \citet{salakhutdinov2008}
combine deep belief networks (DBNs) with Gaussian processes, showing improved performance
over standard GPs with RBF kernels, in the context of semi-supervised learning.  However,
their model is heavily relying on unsupervised pre-training of DBNs, with
the GP component unable to scale beyond a few thousand training points.  Likewise,
\citet{calandra2014manifold} combine a feedforward neural network transformation with a
Gaussian process, showing an ability to learn sharp discontinuities.
However, similar to many other approaches, the resulting model
can only scale to at most a few thousand data points.

In a frequentist setting, \citet{yang2014deep} combine convolutional networks, with parameters
pre-trained on ImageNet, with a scalable Fastfood \citep{le2013fastfood} expansion for the
RBF kernel applied to the final layer.  The resulting method is scalable and flexible, but the
network parameters generally must first be trained separately from the Fastfood features,
and the combined model remains parametric, due to the parametric expansion provided by
Fastfood. Careful attention must still be paid to training procedures, regularization,
and manual calibration of the network architecture.  In a similar manner, \citet{huang2015scalable}
and \citet{snoek2015scalable} have combined deep architectures with parametric Bayesian models.
\citet{huang2015scalable} pursue an unsupervised pre-training procedure using deep autoencoders,
showing improved performance over GPs using standard kernels.  \citet{snoek2015scalable} show promising
performance on Bayesian optimisation tasks, for tuning the parameters of a deep neural network.

Our approach is distinct in that we combine deep feedforward and convolutional architectures with
spectral mixture covariances \citep{wilsonadams2013}, inducing points, Kronecker and Toeplitz
algebra, and local kernel interpolation \citep{wilsonnickisch2015, wdn2015}, to derive
expressive and scalable \emph{closed form} kernels, which can be trained jointly with a unified
supervised objective, as part of a
\emph{non-parametric} Gaussian process framework, without requiring approximate
Bayesian inference.  Moreover, the simple joint learning procedure in our approach 
can be applied in general settings.  Indeed we show that the proposed model outperforms
state of the art stand-alone deep learning architectures and Gaussian processes
with advanced kernel learning procedures on a wide range of datasets, demonstrating
its practical significance.  We achieve scalability while retaining non-parametric model structure by leveraging the
very recent KISS-GP approach~\citep{wilsonnickisch2015} and extensions in \citet{wdn2015} for efficiently representing
kernel functions, to produce scalable deep kernels.

\section{Gaussian Processes}
\label{sec: gps}

We briefly review the predictive equations and marginal likelihood for Gaussian processes (GPs), and 
the associated computational requirements, following the notational conventions in 
\citet{wdn2015}.  See, for example, \citet{rasmussen06} for a comprehensive 
discussion of GPs.

We assume a dataset $\mathcal{D}$ of $n$ input (predictor) vectors
$X = \{\bm{x}_1,\dots,\bm{x}_n\}$, each of dimension $D$,
which index an $n \times 1$ vector of targets
$\bm{y} = (y(\bm{x}_1),\dots,y(\bm{x}_n))^{\top}$.  If
$f(\bm{x}) \sim \mathcal{GP}(\mu,k_{\bm{\gamma}})$, then any
collection of function values $\bm{f}$ has a joint Gaussian
distribution,
\begin{align}
 \bm{f} = f(X) = [f(\bm{x}_1),\dots,f(\bm{x}_n)]^{\top} \sim \mathcal{N}(\bm{\mu},K_{X,X}) \,,  \label{eqn: gpdef}
\end{align}
with a mean vector, $\bm{\mu}_i = \mu(x_i)$, and covariance matrix, $(K_{X,X})_{ij} = k_\bm{\gamma}(\bm{x}_i,\bm{x}_j)$, 
determined from the mean function and covariance kernel of the Gaussian process.
The kernel, $k_{\bm{\gamma}}$, is parametrized by $\bm{\gamma}$.  Assuming additive
Gaussian noise, $y(\bm{x})|f(\bm{x}) \sim \mathcal{N}(y(\bm{x}); f(\bm{x}),\sigma^2)$,
the predictive distribution of the GP evaluated at the
$n_*$ test points indexed by $X_*$, is given by
\begin{align}
 \bm{f}_*|X_*,&X,\bm{y},\bm{\gamma},\sigma^2 \sim \mathcal{N}(\mathbb{E}[\bm{f}_*],\text{cov}(\bm{f}_*)) \,, \label{eqn: fullpred}  \\
 \mathbb{E}[\bm{f}_*] &= \bm{\mu}_{X_*}  + K_{X_*,X}[K_{X,X}+\sigma^2 I]^{-1}\bm{y}\,,   \notag \\
 \text{cov}(\bm{f}_*) &= K_{X_*,X_*} - K_{X_*,X}[K_{X,X}+\sigma^2 I]^{-1}K_{X,X_*} \,.  \notag
\end{align}
$K_{X_*,X}$, for example, is an $n_* \times n$ matrix of covariances between
the GP evaluated at $X_*$ and $X$.   $\bm{\mu}_{X_*}$ is the $n_* \times 1$ mean vector,
and $K_{X,X}$ is the $n \times n$ covariance
matrix evaluated at training inputs~$X$.  All covariance (kernel) matrices implicitly depend on the kernel 
hyperparameters $\bm{\gamma}$.

GPs with RBF kernels correspond to models which have an infinite basis expansion in a dual space, and have
compelling theoretical properties: these models are universal approximators, and have prior support to within an arbitrarily small epsilon band of
any continuous function \citep{micchelli2006universal}.
Indeed the properties of the distribution over functions induced by a Gaussian process are controlled by the kernel function.
For example, the popular RBF kernel,
\begin{align}
k_{\text{RBF}}(\bm{x},\bm{x}') = \exp (-\frac{1}{2} ||\bm{x}-\bm{x}' || / \ell^2)  \label{eqn: rbfkernel}
\end{align}
encodes the inductive bias that function values closer together in the input space, in the Euclidean sense, are more correlated.   The complexity of the functions in the
input space is determined by the interpretable length-scale hyperparameter $\ell$.  Shorter length-scales correspond to functions which vary more rapidly
with the inputs $\bm{x}$.

The structure of our data is discovered through learning interpretable kernel hyperparameters.  The marginal likelihood of the targets $\bm{y}$, the probability of the data conditioned only on kernel hyperparameters $\bm{\gamma}$, provides a principled probabilistic framework for kernel learning:
\begin{equation}
 \log p(\bm{y} | \bm{\gamma}, X) \propto -[\bm{y}^{\top}(K_{\bm{\gamma}}+\sigma^2 I)^{-1}\bm{y} + \log|K_{\bm{\gamma}} + \sigma^2 I|]\,,  \label{eqn: mlikeli}
\end{equation}
where we have used $K_{\bm{\gamma}}$ as shorthand for $K_{X,X}$ given~$\bm{\gamma}$.  Note that the expression for the log marginal likelihood in
Eq.~\eqref{eqn: mlikeli} pleasingly separates into automatically calibrated model fit and complexity terms \citep{rasmussen01}.
Kernel learning can be achieved by optimizing Eq.~\eqref{eqn: mlikeli} with respect to $\bm{\gamma}$.

The computational bottleneck for inference is solving the linear system
$(K_{X,X}+\sigma^2 I)^{-1}\bm{y}$, and for kernel learning is computing
the log determinant $\log|K_{X,X}+ \sigma^2 I|$ in the marginal likelihood.  
The standard approach is to compute the Cholesky decomposition of the
$n \times n$ matrix $K_{X,X}$, which
requires $\mathcal{O}(n^3)$ operations and $\mathcal{O}(n^2)$ storage.
After inference is complete, the predictive mean costs $\mathcal{O}(n)$,
and the predictive variance costs $\mathcal{O}(n^2)$, per test point
$\bm{x}_*$.

\section{Deep Kernel Learning}
\label{sec: dkl}

In this section we show how we can contruct kernels which encapsulate the expressive power of deep
architectures, and how to learn
the properties of these kernels as part of a scalable probabilistic
Gaussian process framework.

Specifically, starting from a base kernel $k(\bm{x}_i,\bm{x}_j | \bm{\theta})$ with hyperparameters
$\bm{\theta}$, we transform the inputs (predictors) $\bm{x}$ as
\begin{align}
k(\bm{x}_i,\bm{x}_j | \bm{\theta}) \to k( g(\bm{x}_i, \bm{w}), g(\bm{x}_j, \bm{w}) | \bm{\theta},\bm{w}) \,,  \label{eqn: deepkernel}
\end{align}
where $g(\bm{x},\bm{w})$ is a non-linear mapping given by a deep architecture, such as a deep
convolutional network, parametrized by weights $\bm{w}$. The popular RBF
kernel (Eq.~\eqref{eqn: rbfkernel}) is a sensible choice of base kernel $k(\bm{x}_i,\bm{x}_j | \bm{\theta})$.
For added flexibility, we also propose to use \emph{spectral mixture} base kernels
\citep{wilsonadams2013}:
\begin{align}
k_{\text{SM}}(\bm{x},\bm{x}' | \bm{\theta}) =   
\sum_{q=1}^Q a_q \frac{|\Sigma_q|^{\frac{1}{2}}}{(2\pi)^{\frac{D}{2}}}
\exp \left(-\frac{1}{2} || \Sigma_q ^{\frac{1}{2}} (\bm{x}-\bm{x}') ||^2 \right )
\cos \langle \bm{x}-\bm{x}', 2\pi \bm{\mu}_q \rangle \,.   \label{eqn: smkernel}
\end{align}
The parameters of the spectral mixture kernel $\bm{\theta} = \{{a}_q, \Sigma_q, \bm{\mu}_q\}$
are mixture weights, bandwidths (inverse length-scales), and frequencies.
The spectral mixture (SM) kernel, which forms an expressive basis for all stationary covariance functions, can
discover quasi-periodic stationary structure with an interpretable and succinct representation,
while the deep learning transformation $g(\bm{x},\bm{w})$ captures non-stationary and hierarchical
structure.

\begin{figure}[t!]
\centering
\includegraphics[scale=0.7]{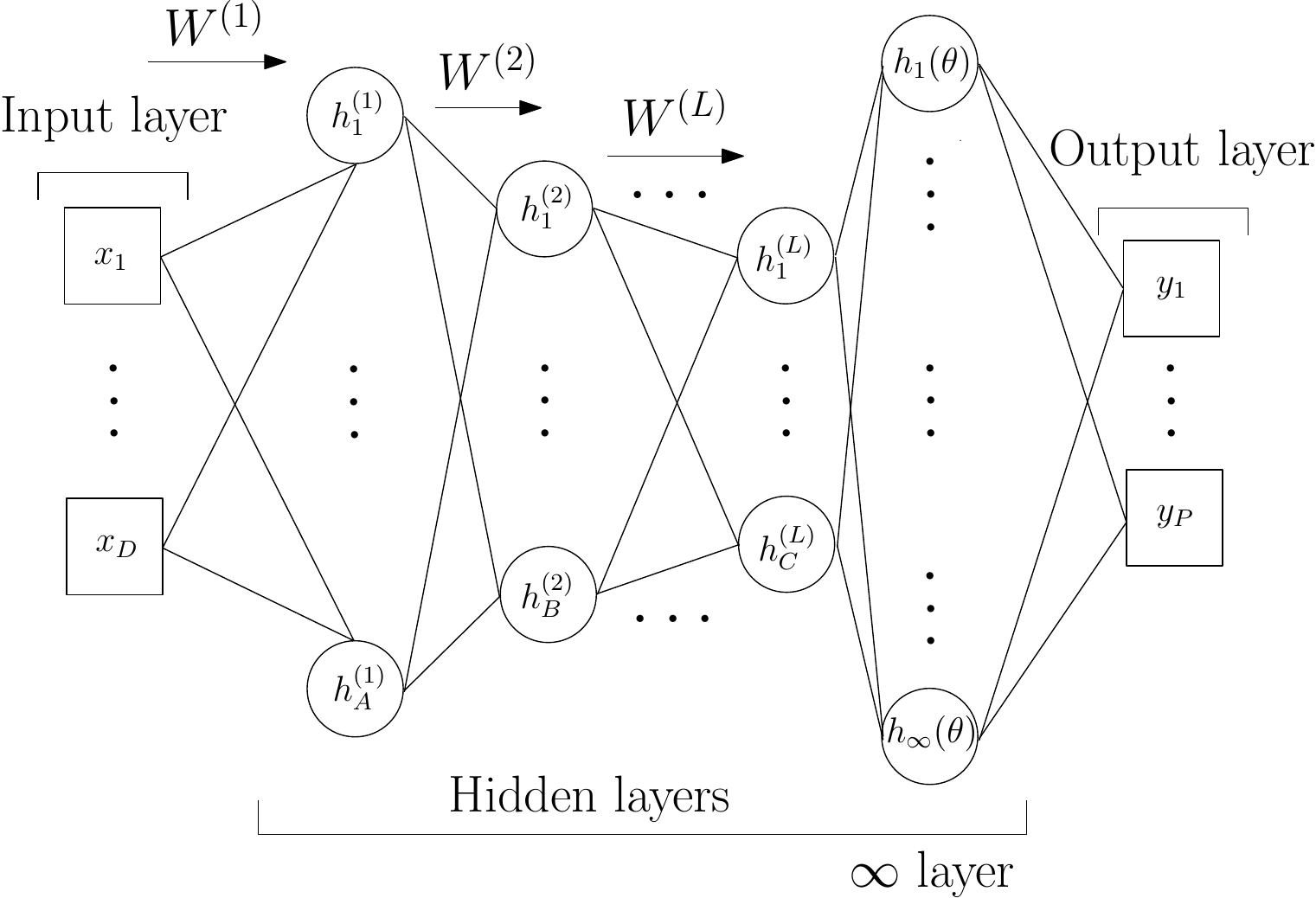}

\caption{\small Deep Kernel Learning: A Gaussian process with a deep kernel maps $D$ dimensional inputs $\bm{x}$ through
$L$ parametric hidden layers 
followed by a hidden layer with  an infinite number of basis functions, with base kernel hyperparameters $\bm{\theta}$. Overall,
a Gaussian process with a deep kernel produces a probabilistic mapping with an infinite number of adaptive basis
functions parametrized by $\bm{\gamma} = \{\bm{w},\bm{\theta}\}$.  All parameters $\bm{\gamma}$ are learned through the marginal likelihood
of the Gaussian process.}
\label{fig: klearnfig}
\end{figure}

We use the deep kernel of Eq.~\eqref{eqn: deepkernel} as the covariance function of a
Gaussian process to model data $\mathcal{D} = \{\bm{x}_i, \bm{y}_i\}_{i=1}^{n}$.
Conditioned on all kernel hyperparameters, we can interpret our model as applying
a Gaussian process with base kernel $k_\bm{\theta}$ to the final hidden layer of a
deep network.  Since a GP with (RBF or SM) base kernel $k_\bm{\theta}$ corresponds to an
infinite basis function representation, our network
effectively has a hidden layer with an \emph{infinite} number of hidden units.
The overall model is shown in Figure~\ref{fig: klearnfig}.

We emphasize, however, that we \emph{jointly} learn all deep kernel hyperparameters,
$\bm{\gamma} = \{ \bm{w},\bm{\theta} \}$,
which include $\bm{w}$, the weights of the
network, and $\bm{\theta}$ the parameters of the base kernel, by maximizing the
log \emph{marginal likelihood}~$\mathcal{L}$ of the Gaussian process (see
Eq.~\eqref{eqn: mlikeli}).  Indeed compartmentalizing our model into a
base kernel and deep architecture is for pedagogical clarity.  When applying
a Gaussian process one can use our deep kernel, which operates as a single
unit, as a drop-in replacement for e.g., standard ARD or Mat\'{e}rn kernels \citep{rasmussen06}, since learning
and inference follow the same procedures.

For kernel learning, we use the chain rule to compute derivatives of the log
marginal likelihood with respect to the deep kernel hyperparameters:
\begin{align}
\label{eq:partial}
\frac{\partial \mathcal{L}}{\partial \bm{\theta}} = \frac{\partial \mathcal{L}}{\partial K_{\bm{\gamma}}} \frac{\partial K_{\bm{\gamma}}}{\partial \bm{\theta}}\,,
\quad \frac{\partial \mathcal{L}}{\partial \bm{w}} = \frac{\partial \mathcal{L}}{\partial K_{\bm{\gamma}}} \frac{ \partial K_{\bm{\gamma}}}{\partial g(\bm{x},\bm{w})}
\frac{\partial g(\bm{x},\bm{w})}{\partial \bm{w}} \,. \notag
\end{align}
The implicit derivative of the log marginal likelihood
with respect to our $n \times n$ data covariance matrix $K_{\bm{\gamma}}$
is given by
\begin{align}
\frac{\partial \mathcal{L}}{\partial K_{\bm{\gamma}}} = \frac{1}{2} (K_{\bm{\gamma}}^{-1} \bm{y}\bm{y}^{\top} K_{\bm{\gamma}}^{-1}
- K_{\bm{\gamma}}^{-1}) \,,
\end{align}
where we have absorbed the noise covariance $\sigma^2 I$ into our covariance
matrix, and treat it as part of the base kernel hyperparameters $\bm{\theta}$.
$\frac{\partial K_{\bm{\gamma}}}{\partial \bm{\theta}}$ are the derivatives of
the deep kernel with respect to the base kernel hyperparameters (such as length-scale),
conditioned on the fixed transformation of the inputs $g(\bm{x},\bm{w})$.  Similarly,
$\frac{ \partial K_{\bm{\gamma}}}{\partial g(\bm{x},\bm{w})}$ are the implicit derivatives
of the deep kernel with respect to $g$, holding $\bm{\theta}$ fixed.  The derivatives
with respect to the weight variables $\frac{\partial g(\bm{x},\bm{w})}{\partial \bm{w}}$ are
computed using standard backpropagation.

For scalability, we replace all instances of $K_{\bm{\gamma}}$ with the KISS-GP
covariance matrix \citep{wilsonnickisch2015, wdn2015}
\begin{align}
K_{\bm{\gamma}} \approx M K^{\text{deep}}_{U,U} M^{\top} := K_{\text{KISS}} \,,
\end{align}
where $M$ is a sparse matrix of interpolation weights, containing only
$4$ non-zero entries per row for local cubic interpolation, and $K_{U,U}$
is a covariance matrix created from our deep kernel, evaluated over $m$ latent inducing
points $U = [\mathbf{u}_i]_{i=1 \dots m}$.
We place the inducing points over a regular multidimensional lattice, and
exploit the resulting decomposition of $K_{U,U}$ into a Kronecker product of
Toeplitz matrices for extremely fast matrix vector multiplications (MVMs), without requiring
any grid structure in the data inputs~$X$ or the transformed inputs $g(\bm{x},\bm{w})$.
Because KISS-GP operates by creating an approximate kernel which admits
fast computations, and is independent from a specific inference and learning procedure,
we can view the KISS approximation applied to our deep kernels as a \emph{stand-alone}
kernel, $k(\bm{x},\bm{z}) = \bm{m}_\bm{x}^{\top} K^{\text{deep}}_{U,U} \bm{m}_\bm{z}$, which can be combined with Gaussian processes or other kernel machines for
scalable learning.

\newgeometry{margin=0.4in}

\begin{table*}[tbhp]
\vspace{-0.2in}
  \centering
  \caption{\small Comparative RMSE performance and runtime on UCI regression datasets, with $n$ training points and $d$ the input dimensions. The results are averaged over $5$ equal partitions (90\% train, 10\% test) of the data $\pm$ one standard deviation.  The {\it best} denotes the best-performing kernel according to \cite{yang2015carte} (note that often the best performing kernel is GP-SM). Following \cite{yang2015carte}, as exact Gaussian processes are intractable on the large data used here, the Fastfood finite basis function expansions are used for approximation in GP (RBF, SM, Best).  We verified on datasets with $n < 10,000$ that exact GPs with RBF kernels provide comparable performance to the Fastfood expansions.
  For datasets with $n < 6,000$ we used a fully-connected DNN with a [$d$-1000-500-50-2] architecture, and for $n > 6000$ we used a [$d$-1000-1000-500-50-2] architecture.  We consider scalable deep kernel learning (DKL) with RBF and SM base kernels.
  For the SM base kernel, we set $Q=4$ for datasets with $n < 10,000$ training instances, and use $Q=6$ for larger datasets.}
  \fontsize{8pt}{0.9em}\selectfont
\begin{tabular}{@{}L{42pt}  R{40pt}  R{20pt}  r  r  r  r  r  r r r r} \cmidrule[\heavyrulewidth](r){1-9} \cmidrule[\heavyrulewidth](l){10-12} %\toprule
    \multirow{3}{*}{Datasets} & \multirow{3}{*}{n} &  \multirow{3}{*}{d} & \multicolumn{6}{c}{RMSE} & \multicolumn{3}{c}{Runtime(s)} \\\cmidrule(lr){4-9} \cmidrule(l){10-12}
    & & & \multicolumn{3}{c}{GP}  & \multirow{2}{*}{DNN} & \multicolumn{2}{c}{DKL} & \multirow{2}{*}{DNN} & \multicolumn{2}{c}{DKL} \\ \cmidrule(l){4-6}\cmidrule(r){8-9}\cmidrule(r){11-12}
    & & & RBF & SM & best & & RBF & SM & & RBF & SM \\
\cmidrule(r){1-9}\cmidrule(l){10-12}
    Gas & 2,565 & 128 & 0.21$\pm$0.07 & 0.14$\pm$0.08 & 0.12$\pm$0.07 &
                            0.11$\pm$0.05 & 0.11$\pm$0.05 & {\bf 0.09$\pm$0.06} &
                            7.43 & 7.80 & 10.52 \\

    Skillcraft & 3,338 & 19 & 1.26$\pm$3.14 & 0.25$\pm$0.02 & 0.25$\pm$0.02 &
                            {\bf 0.25$\pm$0.00} & {\bf 0.25$\pm$0.00} & {\bf 0.25$\pm$0.00} &
                            15.79 & 15.91 & 17.08 \\

    SML & 4,137 & 26  & 6.94$\pm$0.51 & 0.27$\pm$0.03 & 0.26$\pm$0.04 &
                    0.25$\pm$0.02 & 0.24$\pm$0.01 & {\bf 0.23$\pm$0.01} &
                    1.09 & 1.48 & 1.92 \\

    Parkinsons & 5,875 & 20 & 3.94$\pm$1.31 & {\bf 0.00$\pm$0.00} & {\bf 0.00$\pm$0.00} &
                          0.31$\pm$0.04 & 0.29$\pm$0.04 & 0.29$\pm$0.04 &
                          3.21 & 3.44 & 6.49 \\

    Pumadyn & 8,192 & 32 & 1.00$\pm$0.00 & 0.21$\pm$0.00 & {\bf 0.20$\pm$0.00} &
                        0.25$\pm$0.02 & 0.24$\pm$0.02 & 0.23$\pm$0.02 &
                        7.50 & 7.88 & 9.77 \\

    PoleTele & 15,000 & 26 & 12.6$\pm$0.3 & 5.40$\pm$0.3 & 4.30$\pm$0.2 &
                             3.42$\pm$0.05 &  3.28$\pm$0.04 & {\bf 3.11$\pm$0.07} &
                              8.02 & 8.27 & 26.95 \\

    Elevators  &  16,599 & 18 & 0.12$\pm$0.00 & 0.090$\pm$0.001 & 0.089$\pm$0.002 &
                              0.099$\pm$0.001 & {\bf 0.084$\pm$0.002} & {\bf 0.084$\pm$0.002} &
                              8.91 & 9.16 & 11.77 \\

    Kin40k & 40,000 & 8  & 0.34$\pm$0.01 & 0.19$\pm$0.02 & 0.06$\pm$0.00 &
                         0.11$\pm$0.01 & 0.05$\pm$0.00 & {\bf 0.03$\pm$0.01} &
                         19.82 & 20.73 & 24.99 \\

    Protein & 45,730 & 9 & 1.64$\pm$1.66 & 0.50$\pm$0.02 & 0.47$\pm$0.01 &
                         0.49$\pm$0.01 & 0.46$\pm$0.01 & {\bf 0.43$\pm$0.01} &
                         142.8 & 154.8 & 144.2 \\

    KEGG & 48,827 & 22  & 0.33$\pm$0.17 & 0.12$\pm$0.01 & 0.12$\pm$0.01 &
                         0.12$\pm$0.01 & 0.11$\pm$0.00 & {\bf 0.10$\pm$0.01} &
                         31.31 & 34.23 & 61.01 \\

    CTslice & 53,500 & 385 & 7.13$\pm$0.11 & 2.21$\pm$0.06 & 0.59$\pm$0.07 &
                           0.41$\pm$0.06 & 0.36$\pm$0.01 & {\bf 0.34$\pm$0.02} &
                           36.38 & 44.28 & 80.44 \\

    KEGGU & 63,608 & 27 & 0.29$\pm$0.12 & 0.12$\pm$0.00 & 0.12$\pm$0.00 &
                           0.12$\pm$0.00 & {\bf 0.11$\pm$0.00} & {\bf 0.11$\pm$0.00} &
                           39.54 & 42.97 & 41.05 \\

    3Droad & 434,874 & 3  & 12.86$\pm$0.09 & 10.34$\pm$0.19 & 9.90$\pm$0.10 &
                         7.36$\pm$0.07 & {\bf 6.91$\pm$0.04} & {\bf 6.91$\pm$0.04} &
                         238.7 & 256.1 & 292.2 \\

    Song & 515,345 & 90 & 0.55$\pm$0.00 & 0.46$\pm$0.00 & 0.45$\pm$0.00 &
                      0.45$\pm$0.02 &  0.44$\pm$0.00 &  {\bf 0.43$\pm$0.01} &
                      517.7 & 538.5 & 589.8 \\

    Buzz & 583,250 & 77  & 0.88$\pm$0.01 & 0.51$\pm$0.01 & 0.51$\pm$0.01 &
                       0.49$\pm$0.00 & 0.48$\pm$0.00 &  {\bf 0.46$\pm$0.01} &
                       486.4 & 523.3 & 769.7 \\

    Electric & 2,049,280 & 11 & 0.230$\pm$0.000 & 0.053$\pm$0.000 & 0.053$\pm$0.000 &
                       0.058$\pm$0.002 & 0.050$\pm$0.002 & {\bf 0.048$\pm$0.002} &
                       3458 & 3542 & 4881 \\
\cmidrule[\heavyrulewidth](r){1-9} \cmidrule[\heavyrulewidth](l){10-12}
\end{tabular}
\label{tab:uci}
\vspace{-10pt}
\end{table*}

\restoregeometry

For inference we solve $K_{\text{KISS}}^{-1} \bm{y}$ using linear conjugate gradients (LCG), an
iterative procedure for solving linear systems which only involves matrix vector multiplications (MVMs).
The number of iterations required for convergence to within machine precision is
$j \ll n$, and in practice $j$ depends on the conditioning of the KISS-GP covariance matrix
rather than the number of training points $n$.  For estimating the log determinant in
the marginal likelihood we follow the approach described in \citet{wilsonnickisch2015} 
with extensions in \citet{wdn2015}.

KISS-GP training scales as $\mathcal{O}(n + h(m))$ (where $h(m)$ is typically close to linear
in $m$), versus conventional scalable GP approaches which require $\mathcal{O}(m^2 n + m^3)$
\citep{quinonero2005unifying}
computations and need $m \ll n$ for tractability, which results in severe deteriorations in
predictive performance.  The ability to have large $m \approx n$ allows KISS-GP to have near-exact
accuracy in its approximation \citep{wilsonnickisch2015}, retaining a non-parametric representation, while providing linear scaling
in $n$ and $\mathcal{O}(1)$ time per test point prediction \citep{wdn2015}.  We empirically demonstrate this
scalability and accuracy in our experiments of section \ref{sec: experiments}.

\section{Experiments}
\label{sec: experiments}

We evaluate the proposed deep kernel learning method on a wide range of regression problems, including a large and diverse collection of regression tasks from the UCI repository (section~\ref{sec:exp-uci}), orientation extraction from face patches (section~\ref{sec:exp-face}), magnitude recovery of handwritten digits (section~\ref{sec:exp-mnist}), and step function recovery (section~\ref{sec: step}).   We show that the proposed algorithm substantially outperforms Gaussian processes with expressive kernel learning approaches, and deep neural networks, without any significant increases in computational overhead.

All experiments were performed on a Linux machine with eight 4.0GHz CPU cores
and 32GB RAM. We implemented DNNs
based on Caffe~\citep{jia2014caffe}, a general deep learning platform, 
and KISS-GP \citep{wilsonnickisch2015,wdn2015} leveraging GPML \citep{rasmussen10gpml} \footnote{\url{www.gaussianprocess.org/gpml}}.

For our deep kernel learning model, we first train a deep neural network using SGD with the squared loss objective,
and rectified linear activation functions.
After the neural network has been pre-trained, a KISS-GP model was fitted using the top-level features of
the DNN model as inputs. Using this pre-training initialization, our joint deep kernel learning (DKL) model of section \ref{sec: dkl} is then trained by optimizing \emph{all} the hyperparameters
$\bm{\gamma}$ of our
deep kernel, by backpropagating derivatives through the marginal likelihood of the Gaussian process (see Eq.~\ref{eq:partial}).

\begin{figure*}[t]
\hspace{0.3in}
  \subfigure
  {\includegraphics[width=0.42\textwidth]{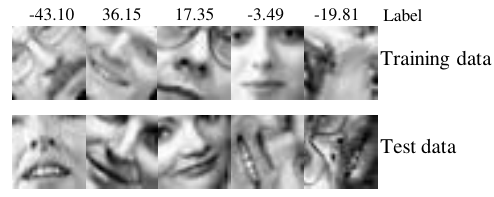}}
\hspace{0.5in}
  \subfigure
  {\includegraphics[width=0.30\textwidth]{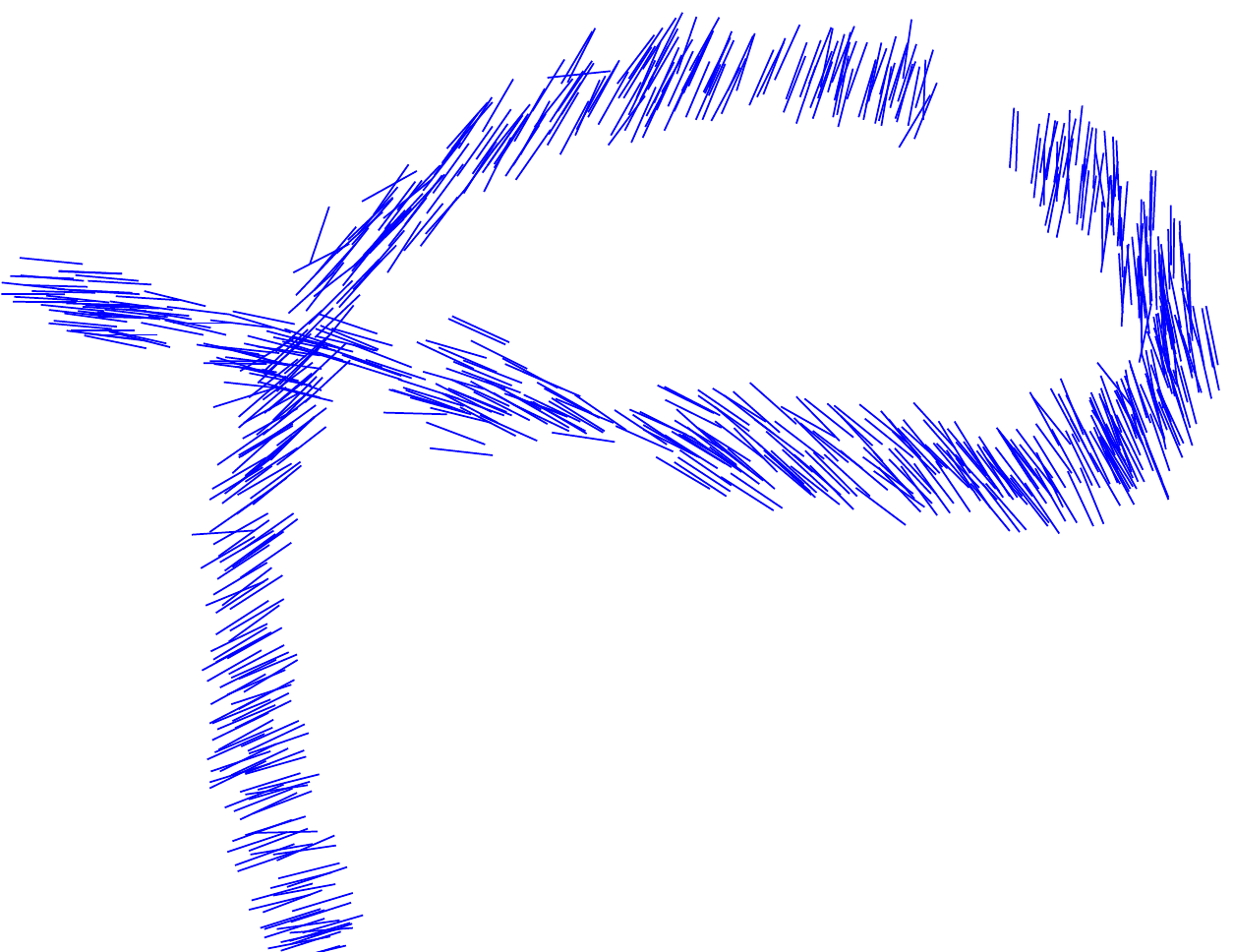}}
  \caption{{\bf Left}: \small Randomly sampled examples of the training and test data.
  {\bf Right}: The two dimensional outputs of the convolutional network on a set of test cases. Each point is shown using a line segment that has the same orientation as the input face.}
\label{fig:face-fea}
\end{figure*}

\subsection{UCI regression tasks}\label{sec:exp-uci}

We consider a large set of UCI regression problems of varying sizes and properties. Table~\ref{tab:uci}
reports test root mean squared error (RMSE) for 1) many scalable Gaussian process kernel learning methods based on Fastfood \citep{yang2015carte}; 2) stand-alone deep neural networks (DNNs); and 3) our proposed combined deep kernel learning (DKL) model using both RBF and SM base kernels.

For smaller datasets, where the number of training examples $n < 6,000$, we used a fully-connected neural network with a d-1000-500-50-2 architecture; for larger datasets we used a d-1000-1000-500-50-2 architecture\footnote{We found [d-1000-1000-500-50] architectures provide a similar level of performance, but scalable Kronecker algebra is most effective if the network maps into $D \leq 5$ dimensional spaces.}.

Table~\ref{tab:uci} shows that
on most of the datasets, our DKL method strongly outperforms not only Gaussian processes with the standard RBF kernel, but also the best-performing kernels selected from a wide range of alternative kernel learning procedures~\citep{yang2015carte}.

We further compared DKL to stand-alone deep neural networks which have the exact same architecture as the DNN component of DKL. By combining KISS-GP with DNNs as part of a joint DKL procedure, we obtain consistently better results than stand-alone deep learning over all 16 datasets.  Moreover, using a spectral mixture base kernel (Eq.~\eqref{eqn: smkernel}) to create a deep kernel provides notable additional performance improvements.
It is interesting to observe that by effectively learning the salient features from raw data, plain DNNs generally achieve competitive performance compared to expressive Gaussian processes.  Combining the complementary advantages of these approaches into scalable deep kernels consistently brings substantial additional performance gains.

We next investigate the runtime of DKL. Table~\ref{tab:uci}, right panel, compares DKL with a stand-alone DNN in terms of runtime for evaluating the objective and derivatives (i.e. one forward and backpropagation pass for DNN; one computation of the marginal likelihood and all relevant derivatives for DNN-KISSGP). We see that in addition to improving accuracy, combining KISS-GP with DNNs for deep kernels introduces only negligible runtime costs: KISS-GP imposes an additional runtime of about 10\% (one order of magnitude less than) the runtime a DNN typically requires.
Overall, these results show the general applicability and practical significance of our scalable DKL approach.

\subsection{Face orientation extraction}
\label{sec:exp-face}

We now consider the task of predicting the orientation of a face extracted from a gray-scale image patch, explored in \citet{salakhutdinov2008}.
We investigate our DKL procedure for efficiently learning meaningful representations from high-dimensional highly-structured image data.

The Olivetti face data set contains ten 64$\times$64 images of forty different people, for $400$ images total.  Following \cite{salakhutdinov2008}, we constructed datasets of 28$\times$28 images by randomly rotating (uniformly from $-90\degree$  to $+90\degree$), cropping, and subsampling the original 400 images. We then randomly select 30 people uniformly and collect their images as training data, while using the images of the remaining 10 people as test data. Figure~\ref{fig:face-fea} shows randomly sampled examples from the training and test data.

\begin{table}[t]
  \centering
  \caption{\small RMSE performance on the Olivetti and MNIST.  For comparison, in the face orientation extraction, we trained DKL on the same amount (12,000) of training instances as with DBN-GP, but used all labels; whereas DBN-GP (as with GP) scaled to only 1,000 labeled images and modeled the remaining data through unsupervised pretraining of DBN. We used RBF base kernel within GPs.\vspace{1mm}}
\begin{tabular}{@{}l  r r r r r }\toprule
 Datasets & GP & DBN-GP & CNN & DKL \\\midrule
 Olivetti & 16.33 & 6.42 & 6.34 & {\bf 6.07} \\
 MNIST & 1.25 & 1.03 & 0.59  & {\bf 0.53} \\
 \bottomrule
\end{tabular}
\label{tab:mnist-face}
\end{table}

\begin{figure*}[t]

\centering
 \subfigure
    {\includegraphics[scale=.4]{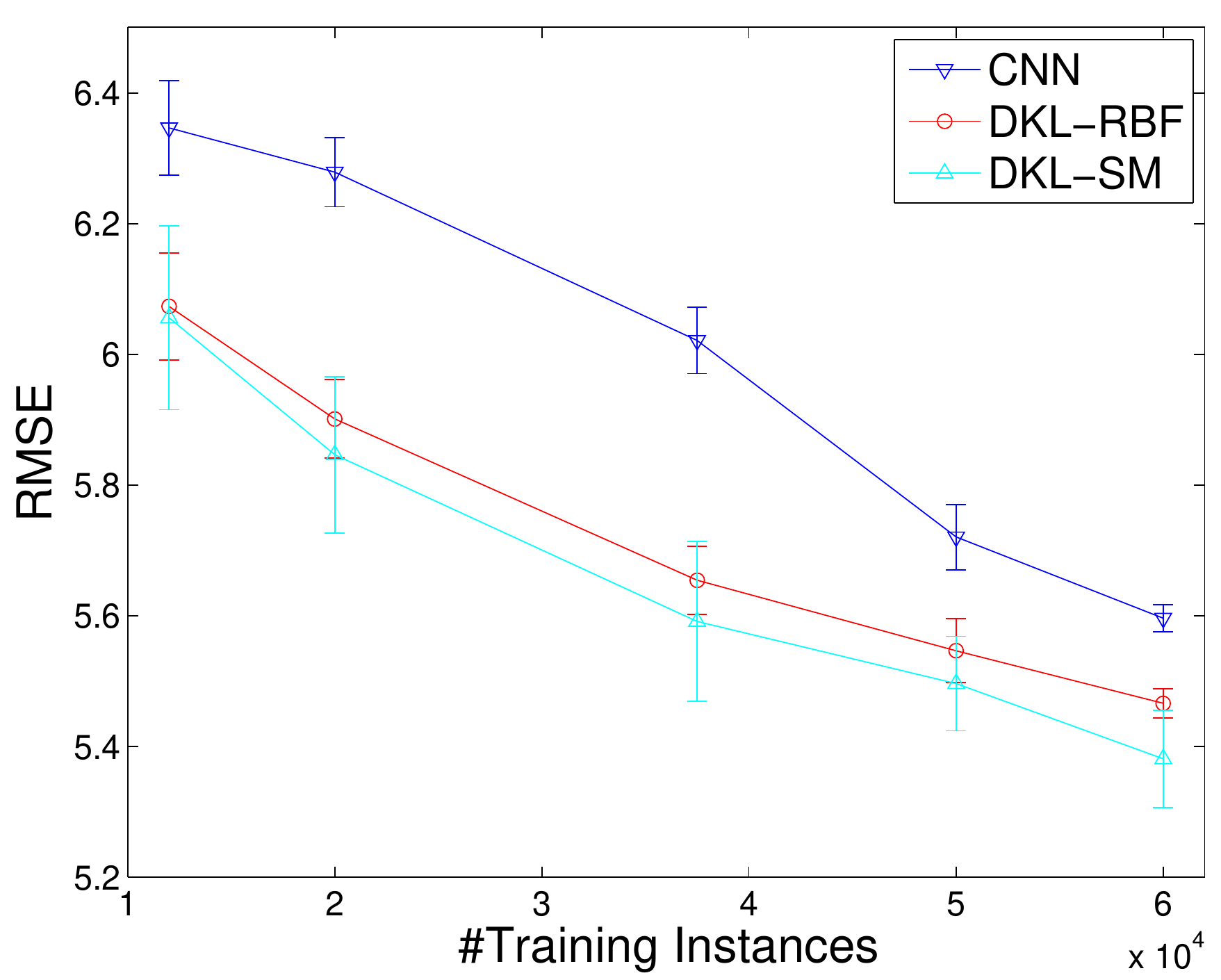}}
  \subfigure
    {\includegraphics[scale=.4]{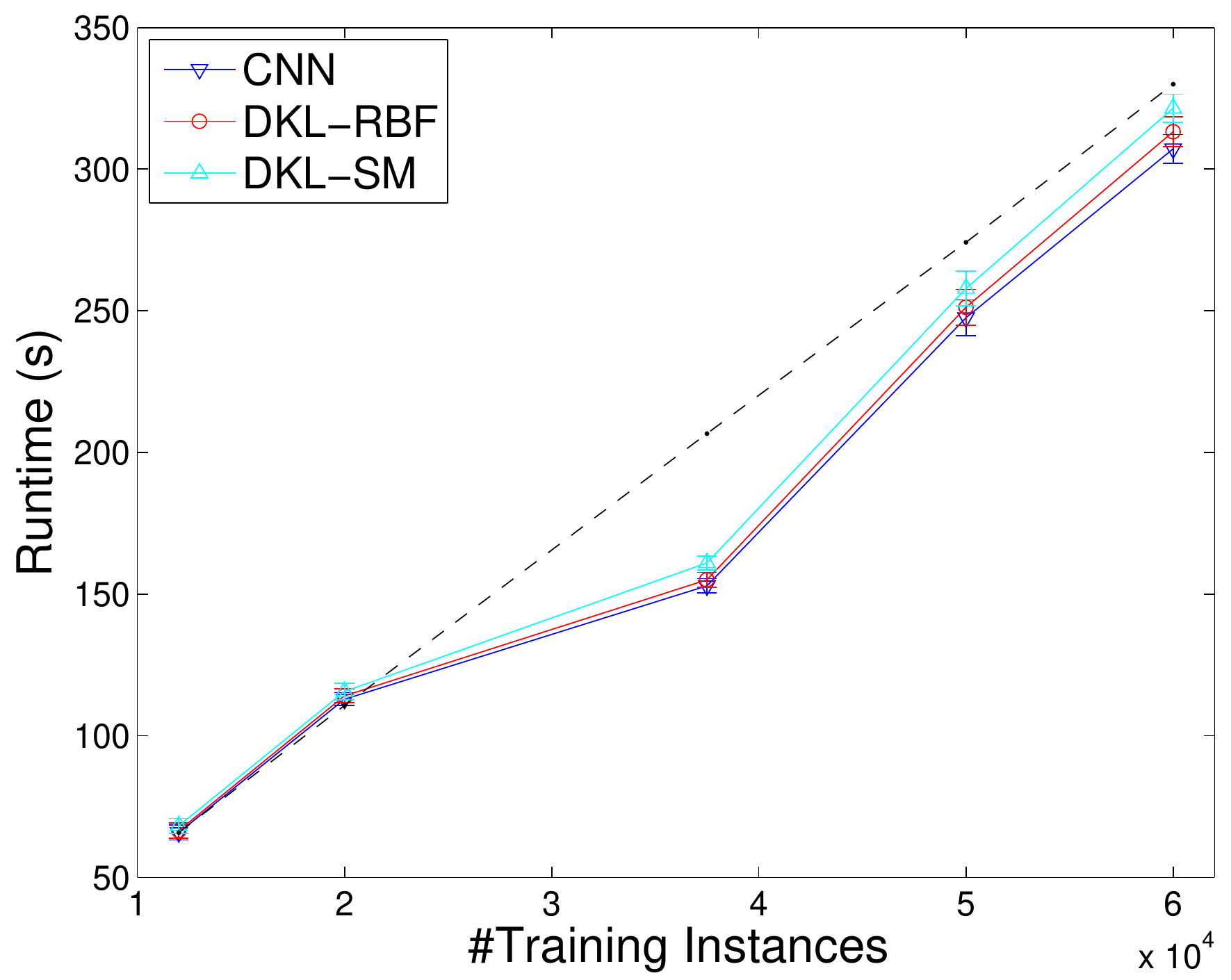}}
  \subfigure
    {\includegraphics[scale=.4]{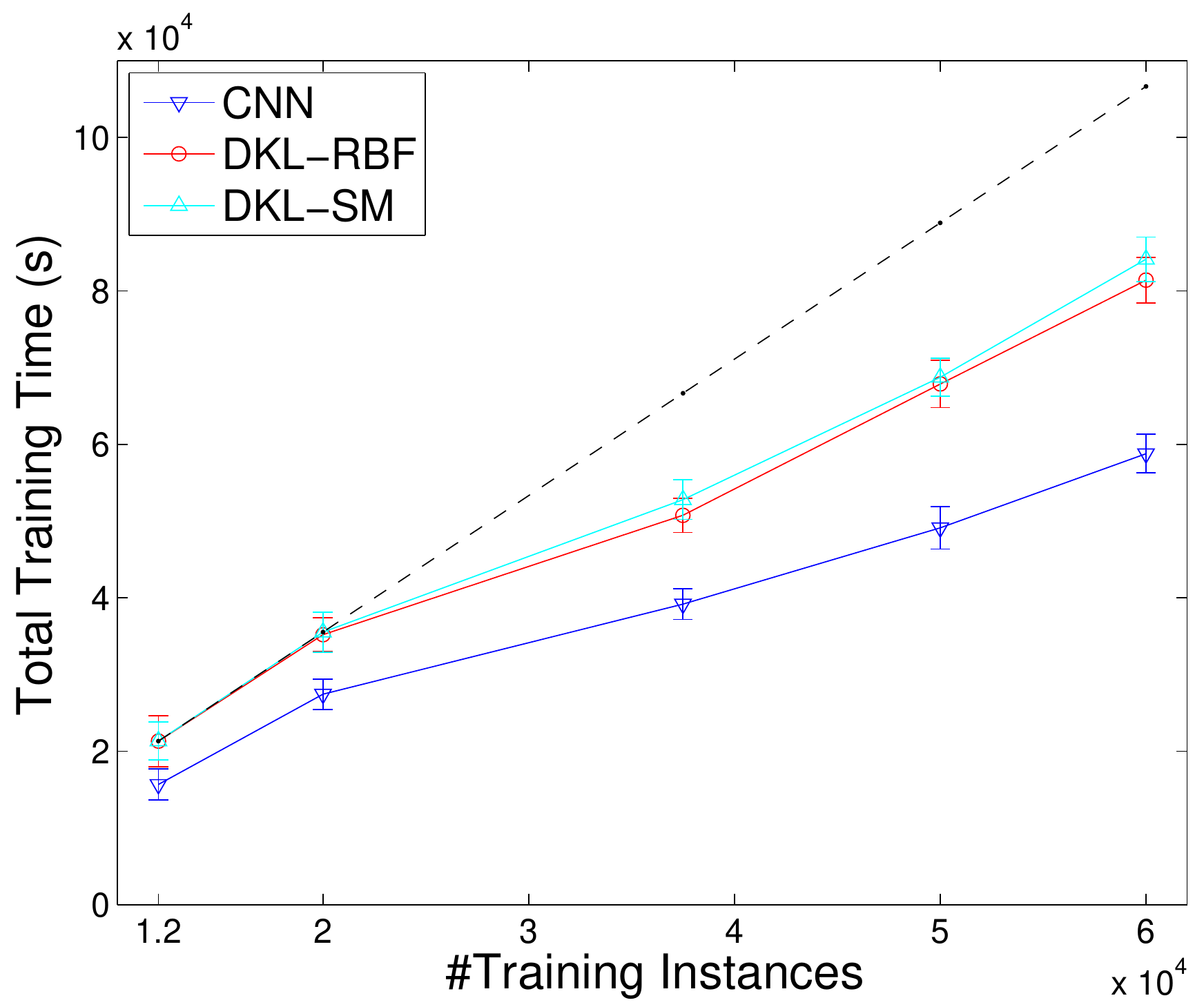}}

  \caption{\small {\bf Left}: RMSE vs. $n$, the number of training examples.
   {\bf Middle}: Runtime vs $n$.
   {\bf Right}: Total training time vs $n$. The dashed line in black indicates a slope of $1$. Convolutional networks are used within DKL. We set $Q=4$ for the SM kernel. }
 \label{fig:rmse-n}

\end{figure*}

For training DKL on the Olivetti face patches we used a convolutional network consisting of 2 convolutional layers followed by 4 fully-connected layers, mapping a face patch to a 2-dimensional feature vector, with a SM base kernel.  We describe this convolutional architecture in detail in the appendix.

Table~\ref{tab:mnist-face} shows the RMSE of the predicted face orientations using four models.
The DBN-GP model, proposed by \cite{salakhutdinov2008}, first extracts features from raw data
using a Deep Belief Network (DBN), and then applies a Gaussian process with an RBF kernel.  However,
their approach could only handle up to a few thousand labelled datapoints, due to the $O(n^3)$ complexity of standard Gaussian processes.
The remaining data were modeled through unsupervised learning of a DBN, leaving the large amount of available labels unused.

Our proposed deep kernel methods, by contrast, scale linearly with the size of training data, and are capable of
directly modeling the full labeled data to accurately recover salient patterns. Figure~\ref{fig:face-fea}, right panel, shows that the deep kernel discovers features essential for orientation prediction, while filtering out irrelevant factors such as identities and scales.

Figure~\ref{fig:rmse-n}, left panel, further validates the benefit of scaling to large data. As more training data are used, our model continues to increase in accuracy. Indeed, it is the large datasets that will provide the greatest opportunities for our model to discover expressive statistical representations.

\begin{figure}[t!]

\centering
\includegraphics[width=0.55\columnwidth]{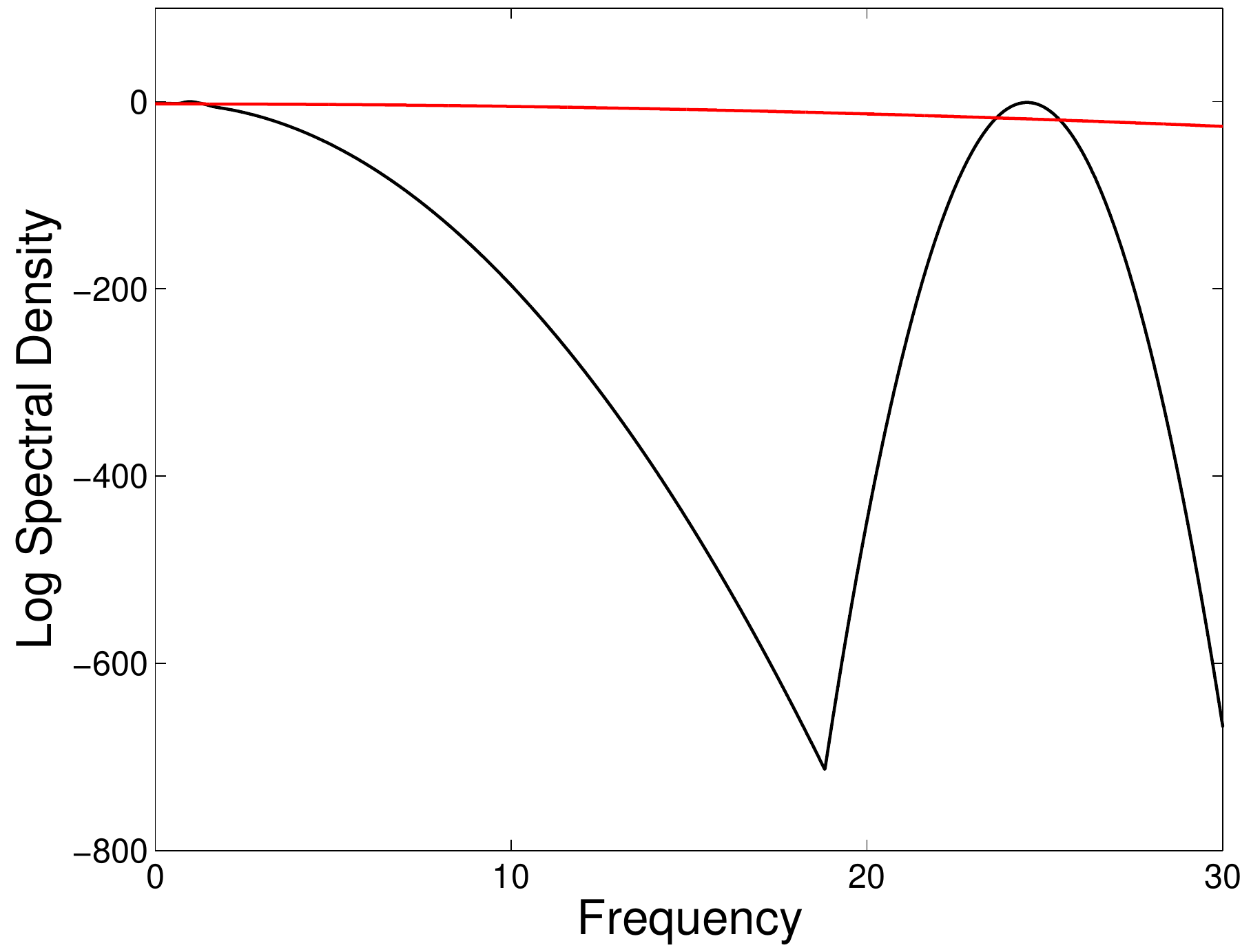}

  \caption{\small The log spectral densities of the DKL-SM and DKL-SE base kernels are in black and red, respectively.}
  \label{fig:kernel-lsd}

\end{figure}

In Figure~\ref{fig:kernel-lsd} we show the spectral density (the Fourier transform) of the \emph{base} kernels learned through our deep kernel learning method. The expressive spectral mixture (SM) kernel discovers a structure with two peaks in the frequency domain.  The RBF kernel is only able to use a single Gaussian in the spectral domain, centred at the origin.  In an attempt to capture the significant mass near a frequency of $25$, the RBF kernel spectral density spreads itself across the whole frequency domain, missing the important local correlations near a frequency $s=0$, thus erroneously discarding much of the network features as white noise, since a broad spectral peak corresponds to a short length-scale.  This result provides intuition for why spectral mixture base kernels generally perform much better than RBF base kernels, despite the flexibility of the deep architecture.

\begin{figure*}[t!]

\centering
  \subfigure
  {\includegraphics[width=0.32\textwidth]{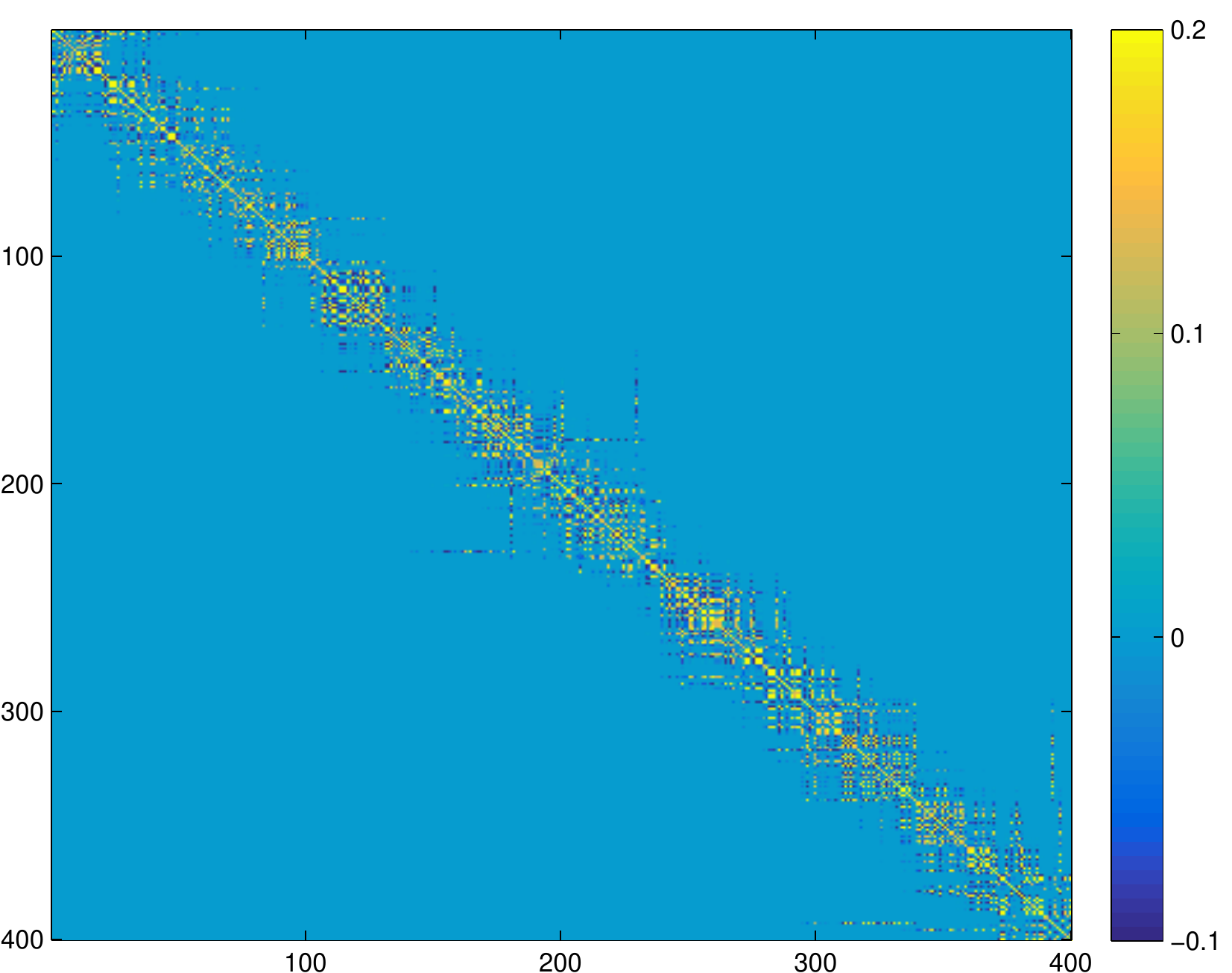}}
  \subfigure
  {\includegraphics[width=0.32\textwidth]{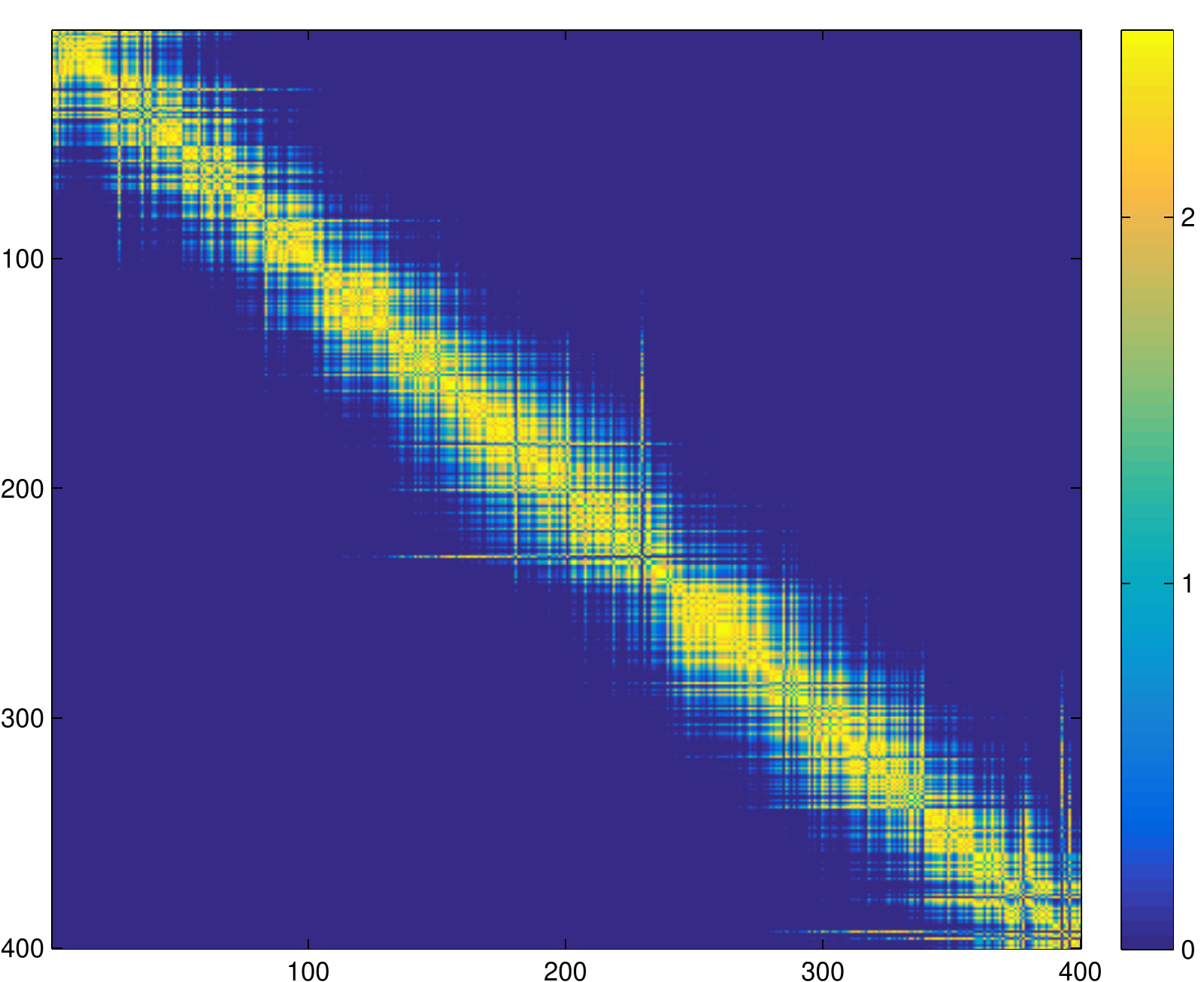}}
  \subfigure
  {\includegraphics[width=0.32\textwidth]{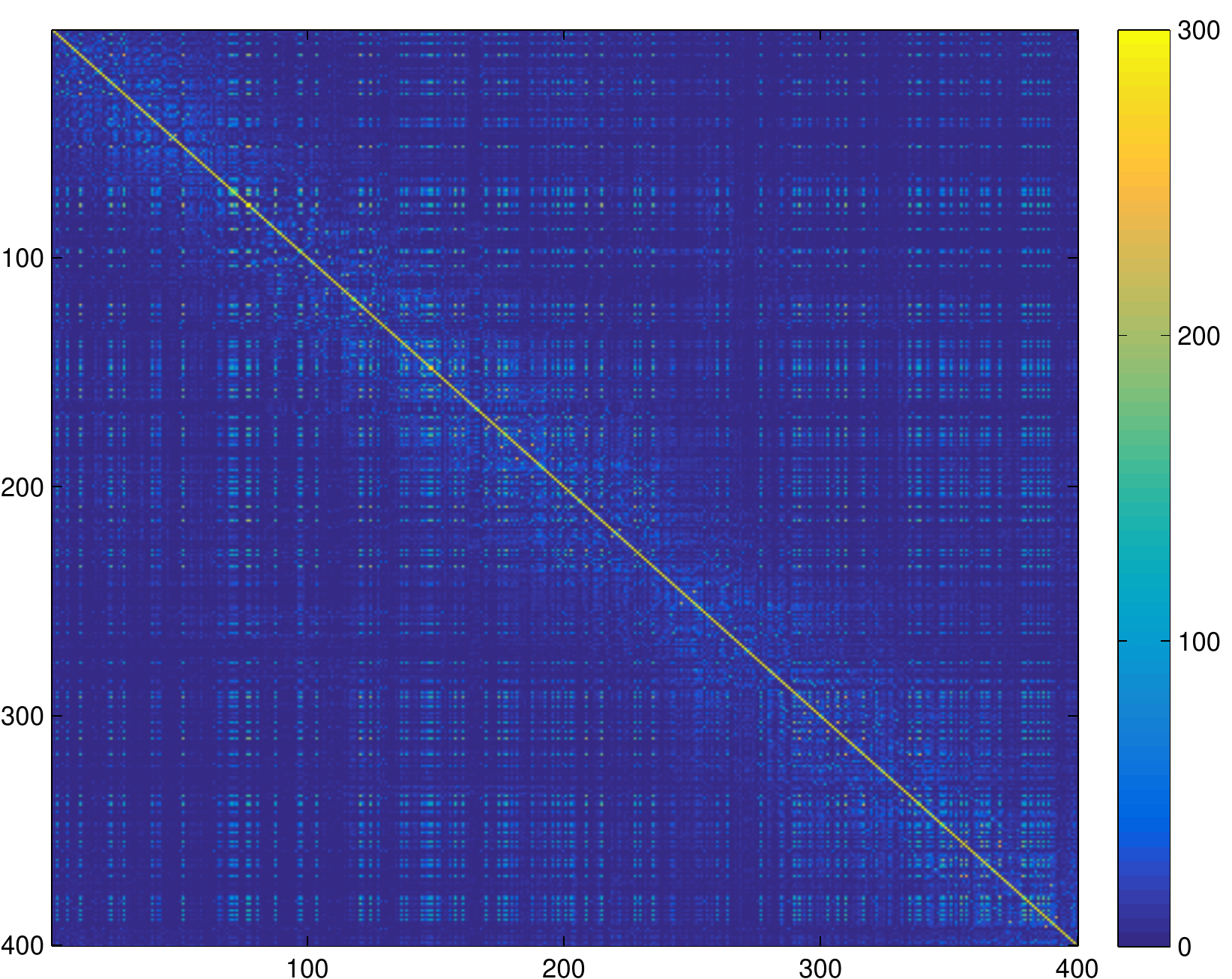}}

  \caption{\small {\bf Left}:
The induced covariance matrix using  DKL-SM kernel
on a set of test cases, where the test samples are ordered according to the {\it orientations} of the input faces. {\bf Middle}: The respective covariance matrix using DKL-RBF kernel. {\bf Right}:
The respective covariance matrix using regular RBF kernel.
The models are trained with $n=12,000$. We set $Q=4$ for the SM base kernel.}

\label{fig:kernel-cov}
\end{figure*}

We further see the benefit of an SM base kernel in Figure \ref{fig:kernel-cov}, where we show the learned covariance matrices constructed from the whole deep kernels (composition of base kernel and deep architecture) for RBF and SM base kernels.  The covariance matrix is evaluated on a set of test inputs, where we randomly sample 400 instances from the test set and sort them in terms of the {\it orientation angles} of the input faces.
We see that the deep kernels with both RBF and SM base kernels discover that faces with similar rotation angles are highly correlated, concentrating their largest entries on the diagonal (i.e., face pairs with similar orientations). Deep kernel learning with an SM base kernel captures these correlations more strongly than the RBF base kernel, which is somewhat more diffuse.

In Figure \ref{fig:kernel-cov}, right panel, we also show the learned covariance matrix for an RBF kernel with a standard Gaussian process applied to the raw data inputs.
We see that the entries are very diffuse.  In essence, through deep kernel learning, we can learn a metric where faces with similar rotation angles are highly correlated, and thus overcome the fundamental limitations of a Euclidean distance metric (used by standard kernels), where similar rotation angles are not particularly correlated, regardless of what hyper-parameters are learned with Euclidean kernels.

We next measure the scalability of our model. Figure~\ref{fig:rmse-n}, middle panel, shows the runtimes in seconds, as a function of training instances, for evaluating the objective and any relevant derivatives. We see that, with the scalable KISS-GP, the joint model achieves a roughly linear asymptotic scaling, with a slope of 1.
In Figure~\ref{fig:rmse-n}, right panel, we show how the total training time (i.e., the time for CNN pre-training plus the time for DKL with CNN architecture joint training) changes with varying the data size $n$. In addition to the linear scaling which is necessary for modeling large data, the added time in combining KISS-GP with CNNs is reasonable, especially considering the gains in performance and expressive power.

\subsection{Digit magnitude extraction}
\label{sec:exp-mnist}
We map images of handwritten digits to a single real-value that is as close as possible to the integer represented by the digit in the image, as in \citet{salakhutdinov2008}. The MNIST digit dataset contains 60,000 training data and 10,000 test $28\times 28$ images of ten handwritten digits (0 to 9). We used a convolutional neural network with a similar architecture as the LeNet~\citep{lecun1998gradient} (detailed in the appendix). Table~\ref{tab:mnist-face} shows that a CNN performs considerably better than GP and DBN-GP, and DKL (with CNN architecture) further improves over CNN.

\subsection{Step function recovery}
\label{sec: step}

\begin{figure}[tb]

\centering
\includegraphics[width=0.6\columnwidth]{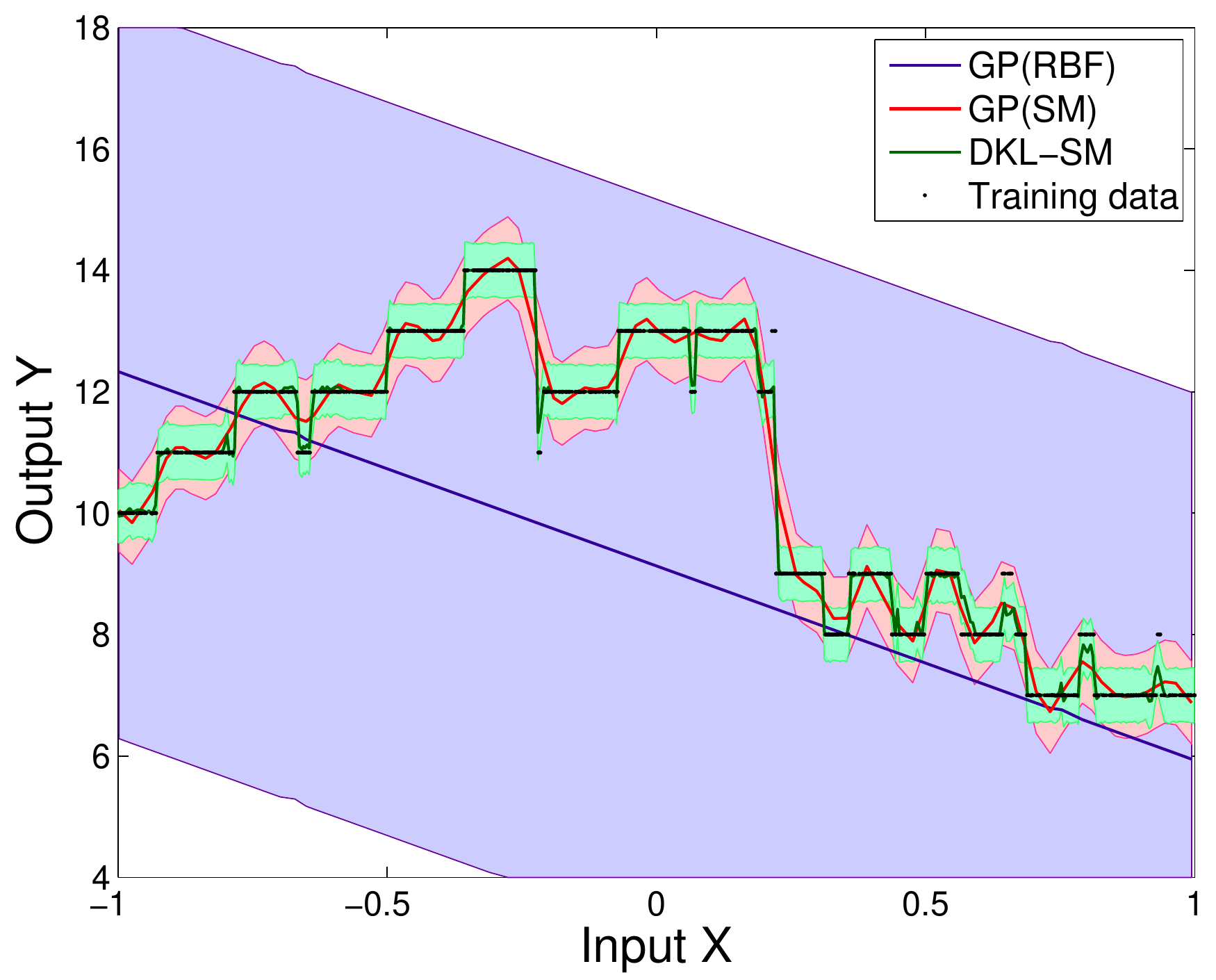}

  \caption{\small Recovering a step function. We show the predictive mean and $95\%$ of the predictive probability mass for regular GPs with RBF and SM kernels, and DKL with SM base kernel. We set $Q=4$ for SM kernels.}
  \label{fig:step}

\end{figure}

We have so far considered RMSE for comparison to alternative methods where posterior predictive distributions are not readily available, or on problems where RMSE has historically been used as a benchmark.  However, an advantage of DKL over stand-alone deep architectures is the ability to naturally produce a posterior predictive distribution, which is especially useful in applications such as reinforcement learning and Bayesian optimisation. 
In Figure~\ref{fig:step}, we consider an example where we use DKL to learn the posterior predictive distribution for a step function with many challenging discontinuities. This problem is particularly difficult for conventional Gaussian process approaches, due to strong smoothness assumptions intrinsic to popular kernels.  

GPs with SM kernels improve upon RBF kernels, but neither can properly adapt to the many sharp changes in covariance structure.  By contrast, the DKL-SM model accurately encodes the discontinuities of the function, and has reasonable uncertainty over the whole domain.

\section{Discussion}
\label{sec: discussion}

We have explored scalable deep kernels, which combine the structural properties of deep architectures with the non-parametric flexibility of kernel methods.   In particular, we transform the inputs of a base kernel with a deep architecture, and then leverage local kernel interpolation, inducing points, and structure exploiting algebra (e.g., Kronecker and Toeplitz methods) for a scalable kernel representation.   These scalable kernels can then be combined with Gaussian process inference and learning procedures for $\mathcal{O}(n)$ training and $\mathcal{O}(1)$ testing time.  Moreover, we use spectral mixture covariances as a base kernel, which provides a significant additional boost in representational power.  Overall, our scalable deep kernels can be used in place of standard kernels, following the same inference and learning procedures, but with benefits in expressive power and efficiency.   We show on a wide range of experiments the general applicability and practical significance of our approach, consistently outperforming scalable GPs with expressive kernels, and stand-alone DNNs.

A major challenge in developing expressive kernel learning approaches is the Euclidean and absolute distance based metrics which are pervasive in most families of kernel functions, such as the ARD and Mat\'{e}rn kernels.  Indeed, although intuitive in some cases, one cannot expect Euclidean and absolute distance as measures of similarity to be generally applicable, and they are especially problematic in high dimensional input spaces \citep{aggarwal2001surprising}.  Modern approaches attempt to learn a flexible parametric family, for example, through weighted combinations of known kernels \citep[e.g.,][]{gonen2011}, but are still fundamentally limited to these standard notions of distance.
As we have seen in the Olivetti faces examples, our approach allows for the whole functional form of the metric to be learned in a flexible manner, through expressive transformations of the input space.  We expect such metric learning to be particularly valuable in high dimensional classification problems, which we view as a promising direction for future research. We hope that this work will help bring together research on neural networks and kernel methods, to inspire many new models and unifying perspectives which combine the complementary advantages of these approaches.

{
\bibliographystyle{apalike}
\bibliography{mbibnew}
}

\clearpage{}\newpage{}
\appendix

\section{Appendix}

\subsection{Convolutional network architecture}

Table~\ref{tab:cnn-arch} lists the architecture of the convolutional networks used in the tasks of face orientation extraction (section~5.2) and digit magnitude extraction (section~5.3). The CNN architecture is original from the LeNet \cite{lecun1998gradient} (for digit classification) and adapted to the above tasks with one or two more fully-connected layers for feature transformation.

\begin{table*}[!thp]
  \centering
  \begin{tabular}{|c|c|c|c|c|c|c|c|c|}
     \hline
     % after \\: \hline or \cline{col1-col2} \cline{col3-col4} ...
     Layer & conv1      & pool1      & conv2      & pool2      & full3 & full4 & full5 & full6 \\ \hline\hline
     kernel size & 5$\times$5 & 2$\times$2 & 5$\times$5 & 2$\times$2 & -     &  -    &   -   & - \\
     stride      & 1          & 2          & 1          & 2          & -     &  -    &   -   & - \\
     channel     & 20         & 20         & 50         & 50         & 1000  & 500   & 50    & 2 \\
     \hline
   \end{tabular}
  \caption{The architecture of the convolutional network used in face orientation extraction. The CNN used in the MNIST digit magnitude regression has a similar architecture except that the {\it full3} layer is omitted. Both {\it pool1} and {\it pool2} are max pooling layers. ReLU layer is placed after {\it full3} and {\it full4}.}
  \label{tab:cnn-arch}
\end{table*}

\end{document}